\documentclass[letterpaper, 10 pt, conference]{ieeeconf}  

\IEEEoverridecommandlockouts                              

\makeatletter
\let\NAT@parse\undefined
\makeatother

\usepackage{graphics} 
\usepackage{graphicx}
\usepackage{grffile} 
\usepackage{amsmath} 
\usepackage{amssymb}  
\usepackage{color}
\usepackage[T1]{fontenc}
\usepackage[utf8]{inputenc}
\usepackage{titlesec}
\usepackage{slashbox}

\usepackage{siunitx}
\usepackage{tabularx}
\usepackage{booktabs, makecell}
\usepackage{float}
\usepackage{seqsplit, collcell, ragged2e}
\newcolumntype{Y}{>{\arraybackslash\collectcell\seqsplit}X<{\endcollectcell}}
\usepackage{lipsum}

\titlespacing*{\section}
{0pt}{3.5ex }{2.5ex plus .2ex}
\titlespacing*{\subsection}
{0pt}{2 ex}{2 ex plus .2ex}

\newenvironment{conditions}
  {\par\vspace{\abovedisplayskip}\noindent\begin{tabular}{>{$}l<{$} @{${}={}$} l}}
  {\end{tabular}\par\vspace{\belowdisplayskip}}
\usepackage{bm}
\usepackage{url}

\usepackage[noend]{algpseudocode}
\usepackage{algorithm}
\usepackage{makecell}
\usepackage[square, comma, numbers,sort&compress]{natbib}
\usepackage{multirow}
\let\labelindent\relax
\usepackage{enumitem}
\usepackage[table,dvipsnames]{xcolor}
\usepackage[pagebackref=true,breaklinks=true,colorlinks,bookmarks=false]{hyperref}
\usepackage{graphicx,booktabs,array}

\hypersetup{
linkcolor=BrickRed
,citecolor=Green
,filecolor=Mulberry
,urlcolor=NavyBlue
,menucolor=BrickRed
,runcolor=Mulberry
,linkbordercolor=BrickRed
,citebordercolor=Green
,filebordercolor=Mulberry
,urlbordercolor=NavyBlue
,menubordercolor=BrickRed
,runbordercolor=Mulberry
}

\title{\LARGE \bf
Mobile 3D Printing Robot Simulation with Viscoelastic Fluids
}

\author{\small Uljad Berdica$^{1,2*}$, Yuewei Fu$^{1*}$, Yuchen Liu$^{1*}$, Emmanouil Angelidis$^{3,4*}$, Chen Feng$^{1\dagger}$\\
\url{https://ai4ce.github.io/M3DP-Sim}
\thanks{$^{*}$ equal contribution.}%
\thanks{$^{1}$New York University, Brooklyn, NY 11201, USA
	{\tt\small \{yf1236, yl5680, cfeng\}@nyu.edu}}%
\thanks{$^{2}$New York University Abu Dhabi, UAE
	{\tt\small ub352@nyu.edu}}%
\thanks{$^{3}$Department of Neuromorphic Computing, fortiss - Research Institute of the Free State of Bavaria
    {\tt\small angelidis@fortiss.org}}
\thanks{$^{4}$Chair of Robotics, Artificial Intelligence and Embedded Systems, Technical University of Munich}
\thanks{$^{\dagger}$Chen Feng is the corresponding author.}%

}

\begin{document}

\maketitle
\thispagestyle{empty}
\pagestyle{empty}

\begin{abstract}

The system design and algorithm development of mobile 3D printing robots need a realistic simulation. They require a mobile robot simulation platform to interoperate with a physics-based material simulation for handling interactions between the time-variant deformable 3D printing materials and other simulated rigid bodies in the environment, which is not available for roboticists yet. To bridge this gap and enable the real-time simulation of mobile 3D printing processes, we develop a simulation framework that includes particle-based viscoelastic fluid simulation and particle-to-mesh conversion in the widely adopted Gazebo robotics simulator, avoiding the bottlenecks of traditional additive manufacturing simulation approaches. This framework is the first of its kind  that enables the simulation of robot arms or mobile manipulators together with viscoelastic fluids. The method is tested using various material properties and multiple collaborating robots to demonstrate its simulation ability for the robots to plan and control the printhead trajectories and to visually sense at the same time the printed fluid materials as a free-form mesh. The scalability as a function of available material particles in the simulation was also studied. A simulation with an average of 5 FPS was achieved on a regular desktop computer.

\end{abstract}

\section{Introduction}
Additive manufacturing (AM) is an increasingly popular production paradigm that replaces labour intensive processes with highly integrated, numerically controlled technologies that facilitate high adaptability to changing trends and increasing complexity demands in product design~\cite{paritala2017digital,rivera2020comprehensive}. With the increased use of such technology, several major issues arise like print size limitation, print process duration~\cite{currencefloor} and different materials integration~\cite{oropallo2016ten} along with print quality~\cite{marques2017mobile}. Advances in robotic technologies~\cite{marques2017mobile,lloret2015complex,bock2015future} have catalyzed the development of mobile 3D printing machinery which can overcome the traditional limitations of gantry-based printing and build arbitrarily large structures through coordinated collaborative processes between printer robots \cite{tiryaki2018printing,McPherson_slicer,marques2017mobile}.

AM projects of such scale require a higher degree of planning for pivotal design decisions like required support~\cite{ostrander2019comparative}, reliable material behavior that maintains the printing patterns after extrusion~\cite{rivera2020comprehensive} and error control for the printer robots and printing materials~\cite{oropallo2016ten}. Therefore, simulation tools for mobile 3D printing are of paramount importance since they help address these issues before production. Current work is either focused on the geometric simulation of collaborative 3D printing~\cite{LS3DPteam,McPherson_slicer} or Finite Elements Method (FEM) that balance the trade-offs between the costly meshing~\cite{chen2019investigation,alharbi2020simulation,10.1115/1.4028580} and signed distance field~\cite{10.1007/978-3-319-32098-4_37} computations.The simulations also lack real time visualization of the layer by layer deposition and temporal change in material behavior~\cite{chen2019investigation}. 

\begin{figure}[t]
    \centering
    \includegraphics[width=0.8\columnwidth]{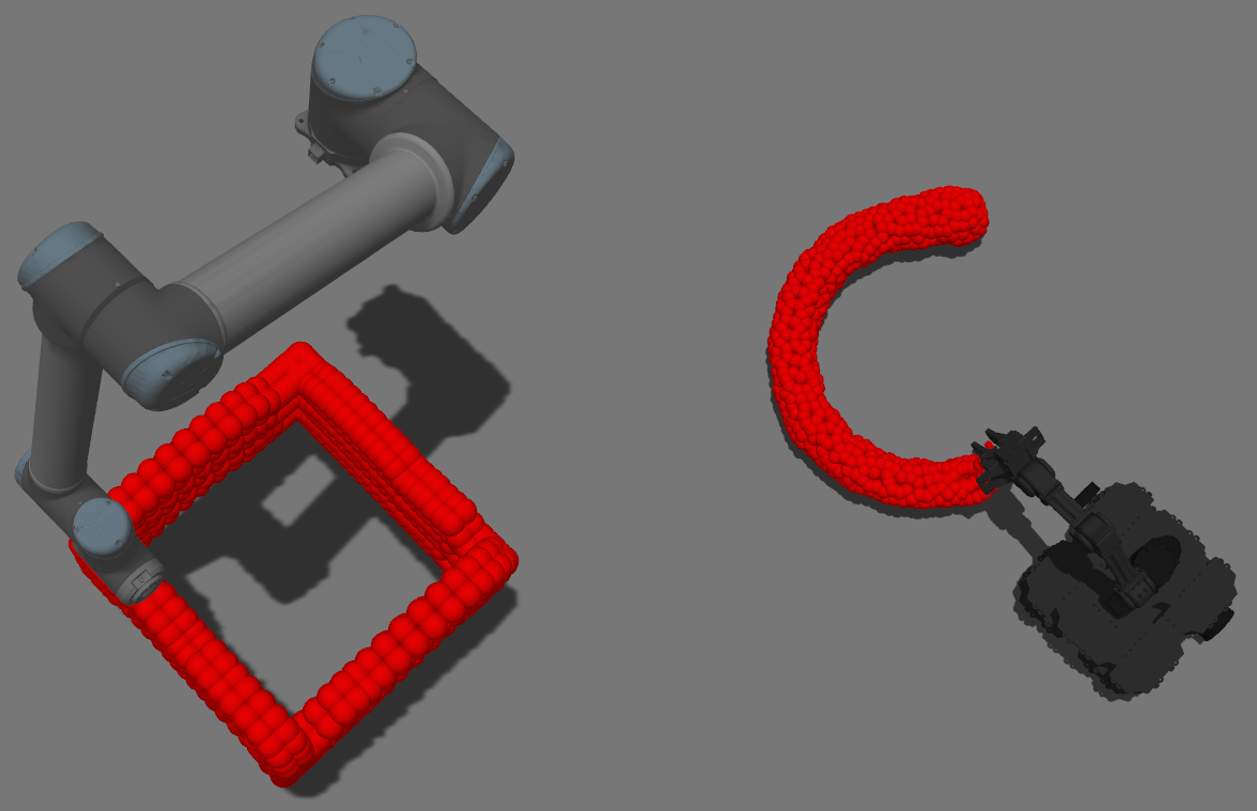}
    \caption{The viscoelastic fluid simulation with a robotic arm printing a squared wall (left) and a Turtlebot3 printing a simple curved geometry (right), each  using  different material properties in the same environment.}
    \label{teaser}
    \vspace{-4mm}
\end{figure}
The contributions of this paper are as follows:
\begin{itemize}
\item[-] The proposal of a robotic AM simulation method that for the first time enables simulated robots to visually perceive, plan, and control viscoelastic fluids, bridging the gap between robotic and AM simulations.
\item[-] The conduction of experiments demonstrating the effectiveness and near real-time simulation frame rates of this framework (Figure \ref{teaser}), implemented as a Gazebo plugin that will be open-sourced for research and education.
\end{itemize}
Note that this paper's goal is to introduce this simulation method for realistic mobile 3D printing robot simulations. We hope the community can take advantage of such new simulation abilities to develop and test new mobile 3D printing robot planning, control, and perception algorithms, which are also the future work of this paper.






\section{Related Work}

\subsection{Mobile 3D printing} The recent works in mobile 3D printing robotics have further established mobile additive manufacturing as a valid solution to the limitations of static AM~\cite{marques2017mobile,LS3DPteam,minibuilders,currencefloor,steckmobile}. \citet{marques2017mobile} presented a mobile fused deposition modelling (FDM) printer along with a circuit design for the remote control of the robot through internet. The work by \textit{The Singapore Center for 3D printing}  utilized a simulation featuring only two robots concurrently working on the same project and was done on OpenRAVE without considering the time evolution of material behavior.

~\citet{McPherson_slicer} propose a "chunk-by-chunk" slicer method to separate the design into multiple parts that can be concurrently printed by a large number of robots. Their simulator can only visualize the geometry of the printing process; the extruded material is represented by a collection of cylinders organized along the interpolated trajectory. This guided the work presented in this paper toward a more granular material physics based approach.

\subsection{AM Simulation and Analysis} Simulations use the predictability of AM processes to check the feasibility of the print,  and control errors that may arise during production~\cite{paritala2017digital,ostrander2019comparative}. 

Finite Elements Analysis of designs was the most used approach both in commercial products like Ansys and COMSOL Multiphysics\textsuperscript{\textregistered}
~\cite{ansys,fortissimomarketplace,alphastarcorporation_2020,comsol} and in academic works~\cite{zhang2020residual, alharbi2020simulation,10.1115/1.4028580,cattenone2019finite}.  \citet{chen2019investigation} pointed out that meshing (an-/isotropic) is the computational bottleneck of FEA methods which require prioritizing resolution for regions of interest.

The works of~\citet{BIEGLER2018158}~and~\citet{lee2019layer} present a challenge that arises from the accumulation of error when printing thin layers on top of one another. Consequently, the physical libraries chosen for this work use a particle-based approach to simulate the extrusion and deposition of the material, bypassing the costly meshing of the entire design and taking into account the change in fluid properties like surface tension and viscosity. A marching cubes methodology similar to that of~\citet{cattenone2019finite} was used to mesh the particles, thus not relying on the computational expenses of meshing to perform the real time simulation.

\subsection{Robotic simulation platform} The process of joining the high fidelity particle-based material simulation with the robot simulation environment is unprecedented in the community and it therefore requires a platform that ensures seamless integration and promotes the use of our tool by the robotics and the AM community. 

The comprehensive reviews of ~\citet{comparison_gaz_vrep} and~\citet{AnalysisComp} cemented the authors' belief that the ease of interfacing with the Robot Operating System (ROS)~\cite{ROS_2009} constitutes a main evaluation criterion in the choice between potential platforms like V-REP, ARGoS, Unity and OpenRave. Surveys in the academic community also reveal that the ratio of participants who use Gazebo as the main tool for simulation is twice larger than that of other listed software in the survey \cite{AnalysisComp}. 

On their introductory paper for Gazebo,~\citet{Gazebo} describe that Gazebo can have significant computational costs when simulating a number of robots exceeding the order of tens. Alternatively, V-REP has shown less CPU usage than Gazebo~\cite{comparison_gaz_vrep} and it features more mesh editing capabilities. While Gazebo exhibits these drawbacks, it was ultimately chosen due to its seamless integration with ROS and the community and the open-source license. The computations overhead is partly made up for by the external libraries used in our simulation and the lack of mesh customization capabilities is accounted for by a new external meshing methodology which offers more relevant options to the context of the AM simulation.

\begin{table}[t]
\caption{The comparison summary of our method vs. related robotic and AM simulation tools}
\label{1234}
\resizebox{\columnwidth}{!}{%
\setlength{\tabcolsep}{4pt}
\setlength{\extrarowheight}{3pt}
\begin{normalsize}
\begin{tabular}{c*{6}{>{\normalsize}c<{\normalsize}c}}

\toprule
    &  \thead{Fluid\\simulation} & \thead{Material\\behavior}& \thead{Multi\\robot~control} & \thead{Real~time\\simulation}  & \thead{Rigid~body\\physics}  & \thead{Mesh\\pipeline} & \\
        \midrule
SOFA\cite{faure2012sofa} &   SPH    & \textbf{+}     &       &  \textbf{+}     &\textbf{+} \\
SPlisHSPlasH\cite{N200BF:2019}  &    SPH    &  \textbf{+}   &      & \textbf{+}  & \\
McPherson et al.\cite{McPherson_slicer}  &      &    &  \textbf{+}    &      &  \textbf{+}\\
V-REP\cite{VREP}  &    &   \textbf{+}&  \textbf{+}    & \textbf{+}    &\textbf{+}\\
Fluid Engine Dev\cite{kim2017fluid}  &   SPH     & \textbf{+}  &    &   \textbf{+}    & &\textbf{+}  \\
\citet{hu2018moving}  &   MPM    &  \textbf{+}   &      &     \textbf{+}   & \textbf{+}  \\
AM in Ansys/COMSOL\cite{ansys,comsol}  &       &  \textbf{+}   &      &       & \textbf{+} & \textbf{+} \\
Cattenone et al.\cite{cattenone2019finite}  &      &  \textbf{+}   &      &       &  & \textbf{+} \\

\textbf{Our tool} &    SPH  & \textbf{+}   & \textbf{+}    & \textbf{+}    & \textbf{+} & \textbf{+}\\

\bottomrule
\end{tabular}
\end{normalsize}
}

\vspace{3mm}
\begin{scriptsize}
The main \textit{Fluid Simulation} method is listed when the tool has a fluid simulation capability that is different from FEM of solid meshes. \textit{Real time} refers to whether the tool has the capacity to render AM processes in real time. \textit{Material behavior} refers to non-fluid material properties, e.g., elasticity.
\end{scriptsize}
\vspace{-3mm}
\end{table}

\subsection{Material Simulation and Meshing Methods} Physics-based material simulation is at the core of the proposed methodology in this paper. The cost efficiency and the relevancy of the methods is just as important as the ability of the user to interact with all the layers of the simulation starting from the solver and constraints to the visual representation. The open-source \textit{Simulation Open Framework Architecture} (SOFA) project uses XML files that allow the user to interface with all the aspects of the simulation before and during run-time~\cite{faure2012sofa}. A similar run-time versatility approach was taken in the meshing process from the location of the simulated particles.

The mathematical models of viscoelastic fluids by~\cite{chang2009particle,gerszewski2009point} directed the attention to \textit{Smooth Particle Hydrodynamics} (SPH) as one of the most viable solutions for a wide range of materials.\textit{SPlisHSPlasH}~\cite{bender2017micropolar,bender2019volume,weiler2016projective,weiler2018physically} was chosen due to its  flexibility in defining the physical properties of the simulation (i.e. resolution, viscosity, stiffness etc) and choice of solvers.

Our~\textit{SPlisHSPlasH} plugin to Gazebo uses an XML file description that encapsulates all these different properties. The method described in~\cite{Akinci2012} was used to sample the rigid boundaries with particles that interact with the fluid particles based on the SPH approach. The forces applied on the boundary particles are subsequently applied on the rigid bodies. The numerical stability of the simulation is guaranteed by the CFL condition, which ensures that the particle with the maximum velocity does not cover a distance longer that twice the particle radius within a timestep, thus preventing particles from overlapping and crossing rigid boundaries. It was found that timesteps at around 1ms were enough to guarantee numerical stability both for the robotics and viscoelastic simulation. The implicit DFSPH method~\cite{Bender2015} was employed for the simulation.
The highly efficient Moving Least Squares Material Point Method (MPM) was also investigated~\cite{hu2018moving,hu2017asynchronous,stomakhin2013material,ram2015material} as an alternative to SPH methods.

Meshing is the final phase of this paper's proposed method as it shows the final simulated print result. \cite{shi2019mixed,wang2017anisotropic,yu2012explicit,orthmann2010topology,wang2012real} were very helpful in designing a meshing pipeline that starts with generating a level set and integrating it with a marching cubes algorithm to extract the surface of free moving particles. Doyub Kim's work on the  \textit{Fluid Engine Development} book~\cite{kim2017fluid} provided a particles-to-object solution that allowed for the generation of an \textit{.obj} file from any given frame and import it to the simulation environment. Thus, this new methodology bypasses the costly meshing bottlenecks pointed out by~\citet{chen2019investigation}.

\begin{figure*}[!t]
    \centering
     \includegraphics[width=0.85\textwidth]{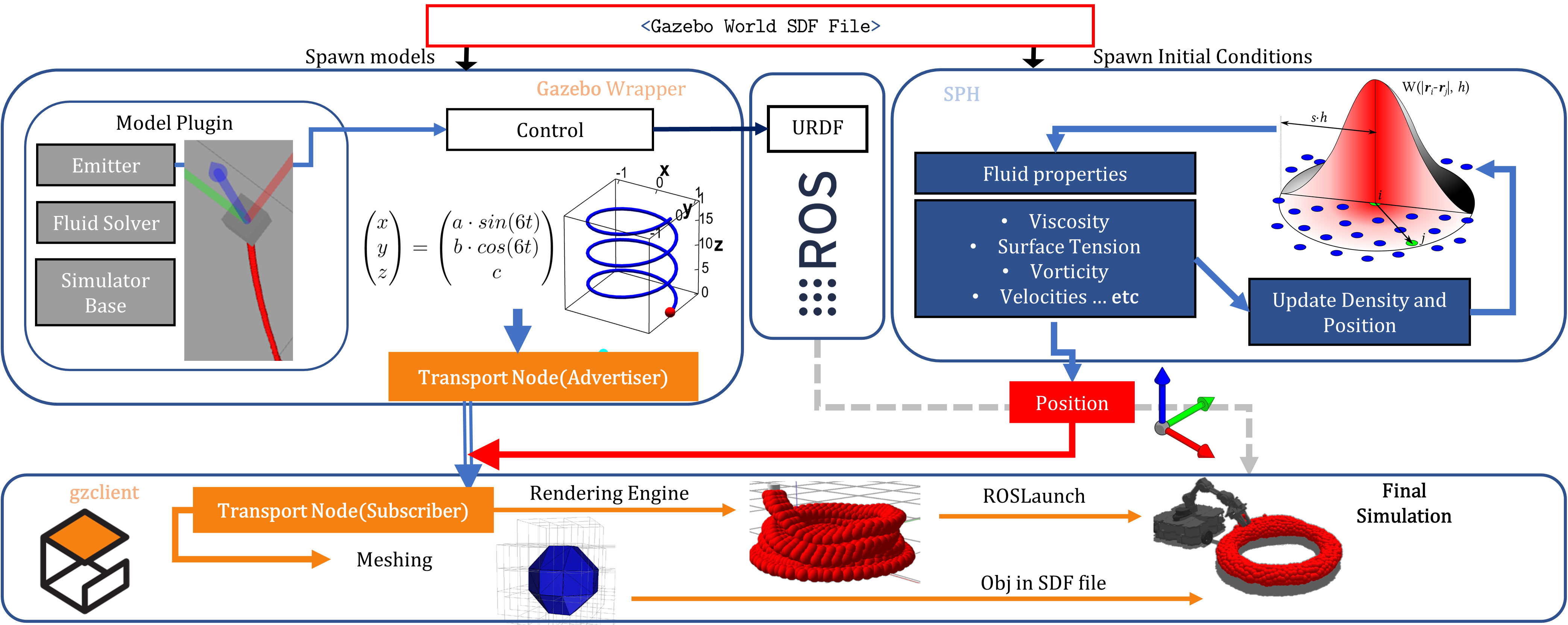}
    \caption{The workflow of our simulation approach. This describes the run from the XML Gazebo world file to the SPH and Gazebo Wrapper code which is then rendered in the Gazebo GUI (gzclient) and meshed in the final object.}
    \label{pipeline}
    \vspace{-3mm}
\end{figure*}

\section{Method}

Figure \ref{pipeline} shows the pipeline of the proposed simulation method. The Gazebo scene file provides the initial conditions which are in turn fed to the SPH library that calculates the new particle positions, as well as the forces applied on the rigid bodies for each step. The particles position is used to render the particles using an object-oriented rendering engine that creates rigid entities of variable size.

\subsection{SPH Overview}
Smoothed particle hydrodynamics (SPH)~\cite{chang2009particle,gerszewski2009point} is at the core of this viscoelastic material simulation. 
The main equations to be solved are the well known equations of continuity for incompressible flow  and the 3D Navier-Stokes equation in Lagrangian coordinates as expressed by~\citet{N200BF:2019}.



\begin{figure}[h!]
    \centering
    \includegraphics[width=0.8\columnwidth]{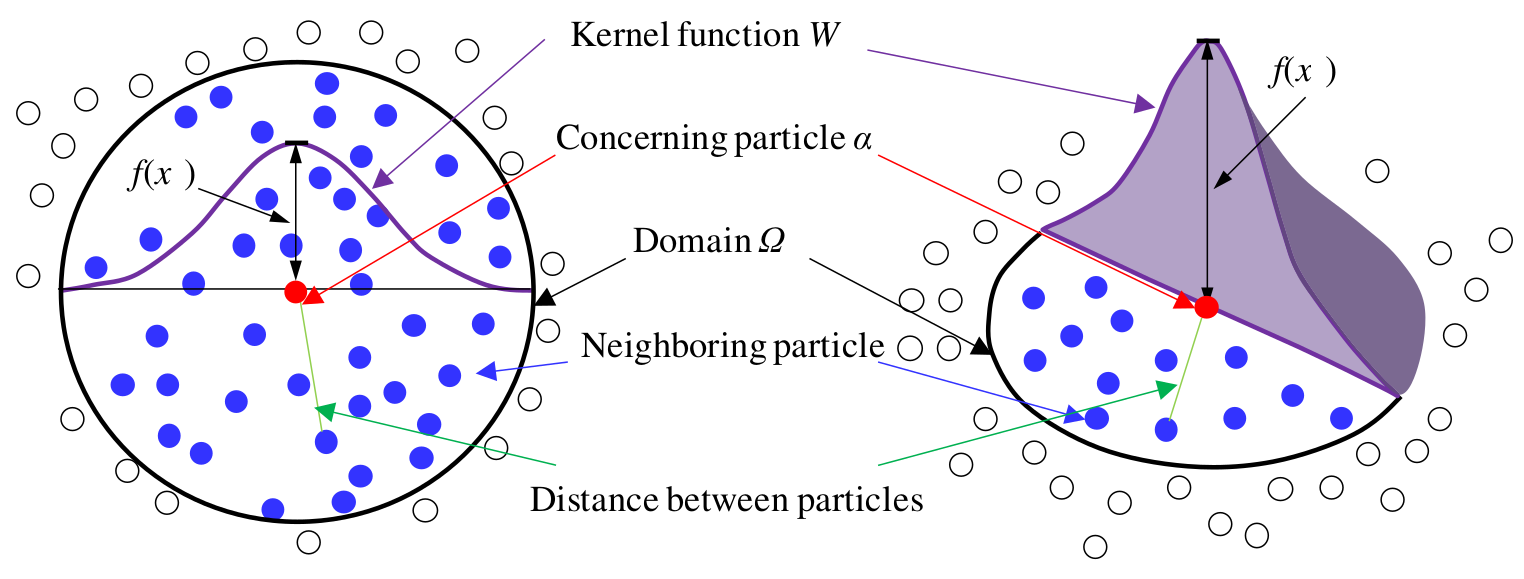}
    \caption{Visualization of the averaging process for a property \textit{f} using a kernel \textit{W(\textbf{r},\textbf{h})}. Source: Adapted from \cite{dai2016sph}.}
    \label{sph1}
    \vspace{-2mm}
\end{figure}

In the SPH formalism, the fluid is discretized to an ensemble of particles representing the physical space, each equipped with a Smoothing Kernel. Each spatial quantity of the particle (f.e. mass, pressure) is computed as a weighted sum of the quantities of its neighbouring particles, within a radius of interaction. The closer the neighbouring particle, is located, the greater the influence it has on the properties of the particle into consideration as illustrated in Figure~\ref{sph1}. More formally let \textit{f} be any particle quantity (such as mass, pressure) to be computed, $\left \langle f(x) \right \rangle$. This function can be rewritten as a the convolution with the Dirac-$\delta$ function over a specified domain $\Omega$: 

\begin{equation}
\left \langle f(x) \right \rangle=\int_{\Omega} f({x}')\delta(x-{x}')d{x}'. \label{avg_val}
\end{equation}
\vspace{-3.5mm}

In SPH, the identity \eqref{sph1} can be approximated by replacing the Dirac-$\delta$ with any kernel $W(\vec{r},h)$ as $h\rightarrow0 $. \textit{h} is the variable radius over which the physical properties are approximated. Equation \eqref{avg_val} can be discretized to compute any property of particle \textit{i} described by a function $f_{i}$ over a domain of a given number of particles.


For a thorough SPH review, we refer the reader to~\cite{N200BF:2019}.

\subsection{Meshing}

\begin{figure}[h!]
    \centering
    \includegraphics[scale=0.5]{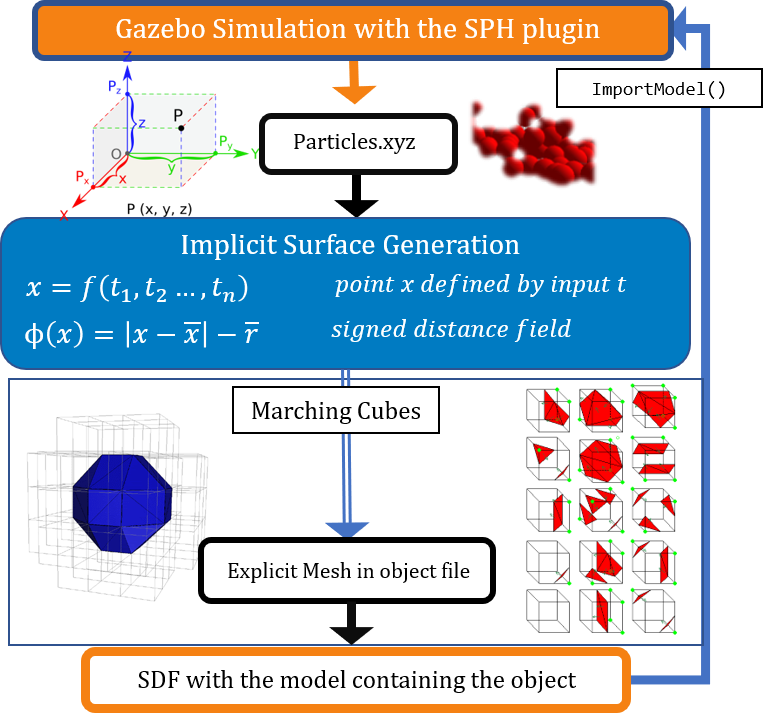}
    \caption{Pipeline of the meshing process from the particle data to the obj file that can be put back in the running simulation. Marching Cubes implementation by~\citet{kim2017fluid}.}
    \label{mesh}
    \vspace{-3mm}
\end{figure}

The meshing process takes the Cartesian coordinates of the fluid model calculated in SPH and creates an implicit surface level set whose boundaries is determined by the points that satisfy an equation $F(x,y,z)$. Through the marching cubes algorithm, a uniform-density polygonal mesh can be extracted from the implicitly defined surface~\cite{lorensen1987marching}. Once the object file is generated, it is packaged in a Simulation Description Format (SDF) file which is then inserted back to the simulation during run time. Figure \ref{mesh} shows the flow of the particle data from the simulation to the explicit triangulation functions and back to the Gazebo environment.

As~\citet{zhu_bridson} explained, a distance function defines the surface that can generate an explicit mesh through the marching cubes algorithm:
\begin{equation}
\phi(x)=\left | x-\overline{x} \right |-\overline{r}, \label{surface}
\end{equation}
\vspace{-5mm}
\begin{conditions}
\overline{r} & weighted average of the particles radii, \\
\overline{x} & weighted average of the neighboring particles position.\\
\end{conditions}

In~\cite{zhu_bridson}, the weight is assigned through:

\vspace{-2mm}
\begin{equation}
w_{i}=\frac{k(\left | x-x_{i} \right |/R)}{\sum_{j}^{}k(\left | x-x_{j} \right |/R)}, \label{weight}
\end{equation}
\vspace{-2mm}
\begin{conditions}
R & radius of the neighborhood around particle $x_{i}$, \\
k & kernel with tails that drop to zero.\\
\end{conditions}

The choice of kernel is non-trivial for special concave regions or small holes since it can lead to the average of the particle being outside the surface~\cite{zhu_bridson}. The simulation of the interaction of material with different densities poses another challenge which can be resolved by the using an anisotropic kernel to capture the densities more accurately at the expense of higher computational costs~\cite{wang2017anisotropic}. The application of Laplacian smoothing and the principal components analysis performed are covered in detail on ~\cite{wang2012real,yu2013reconstructing}. Other faster, less accurate algorithms like spherical and blobby do not have the averaging step Zhu-Bridson and Anisotropic methods have. Through the process shown in Figure \ref{mesh}, the user can choose between the methods based on their simulation needs.

\subsection{Integration}
\vspace{-2mm}
The libraries for the fluid simulation and the meshing of the print results are integrated in the workflow in Figure \ref{pipeline}. The XML file reads the Gazebo world environment configuration and the initial conditions for the fluid simulation. Gazebo computes the positions of the rigid bodies and updates them on the SPH simulator side. This way the robotic simulation is influenced by the presence of the viscoelastic fluid particles. The Gazebo Messages library helps communicate the Gazebo topics through Google protobuf messages in the {\fontfamily{lmtt}\selectfont GazeboGenerator} and  {\fontfamily{lmtt}\selectfont MsgFactory} class. The node connected to the simulation method will keep sending messages including positions of all the particles. The {\fontfamily{lmtt}\selectfont Subscriber} node calls the rendering function through the Object-Oriented Graphics Rendering Engine.

In Gazebo, a particle emitter is attached to a cylindrical model which allows for a simplified representation of the print head attached to the robot. The cylindrical model is controlled by a {\fontfamily{lmtt}\selectfont gazebo::physics::model} plugin. The movement of this model is instrumental in  checking the printing patterns and different levels of curvature and thickness. The rigid boundaries are sampled by particles that behave like the viscoelastic material according to~\cite{Akinci2012} ensuring that interaction forces are applied on the boundary particles and on the rigid bodies. This ensures that the collisions between the robot and the viscoelastic material are handled properly.

\vspace{-4mm}



\section{Experiments}

To demonstrate the validity of the proposed approach, a simulation is used to show how the results are affected by different material properties. Furthermore, the simulation plugin is integrated with various robots in order to connect the viscoelastic material simulation to mobile 3D printing simulations. A performance analysis is carried out to demonstrate the trade-offs of the approach proposed in this paper which allows the user to prioritize based on the objective of their simulation.

\subsection{Simple particle dropping demonstration}
\begin{figure}[h!]
    \centering
    \includegraphics[width=0.65\columnwidth]{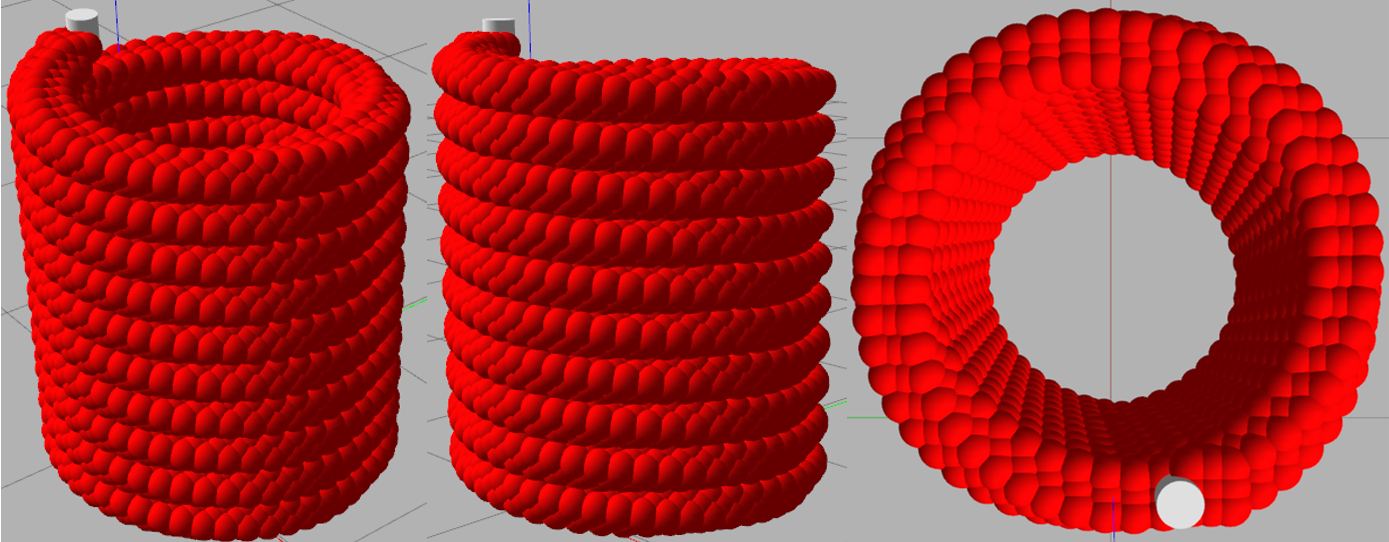}
    \vspace{-2mm}
    \caption{The printing material is rendered into discrete rigid spheres. This does not take into account the viscoelastic properties of the particle system. }
    \label{dropping}
    \vspace{-5mm}
\end{figure}

As seen in the literature review, certain simulations may require little to no realistic material behavior and are focused on the geometrical layout of the layer by layer printing process. Therefore, the first experiment consists in printing a cylindrical wall using only rigid entities under the normal gravity force. Figure \ref{dropping} shows the simulation of a simple block-dropping print with the emitter head represented by a small cylinder.
\vspace{-1mm}

\subsection{Simulating Different Material Behavior}

After having demonstrated the ability to show the layer by layer printing using rigid bodies, this experiments shows the capability of the SPH libraries to direct the rendered entities to behave as a fluid with time dependent viscosity. Viscosity coefficient $\mu$, can be calculated from equation~\eqref{NavStok} as product of kinematic viscosity coefficient $\eta$ and density $\rho$ . 

\begin{figure}[h!]
    \centering
    \includegraphics[width=0.8\columnwidth]{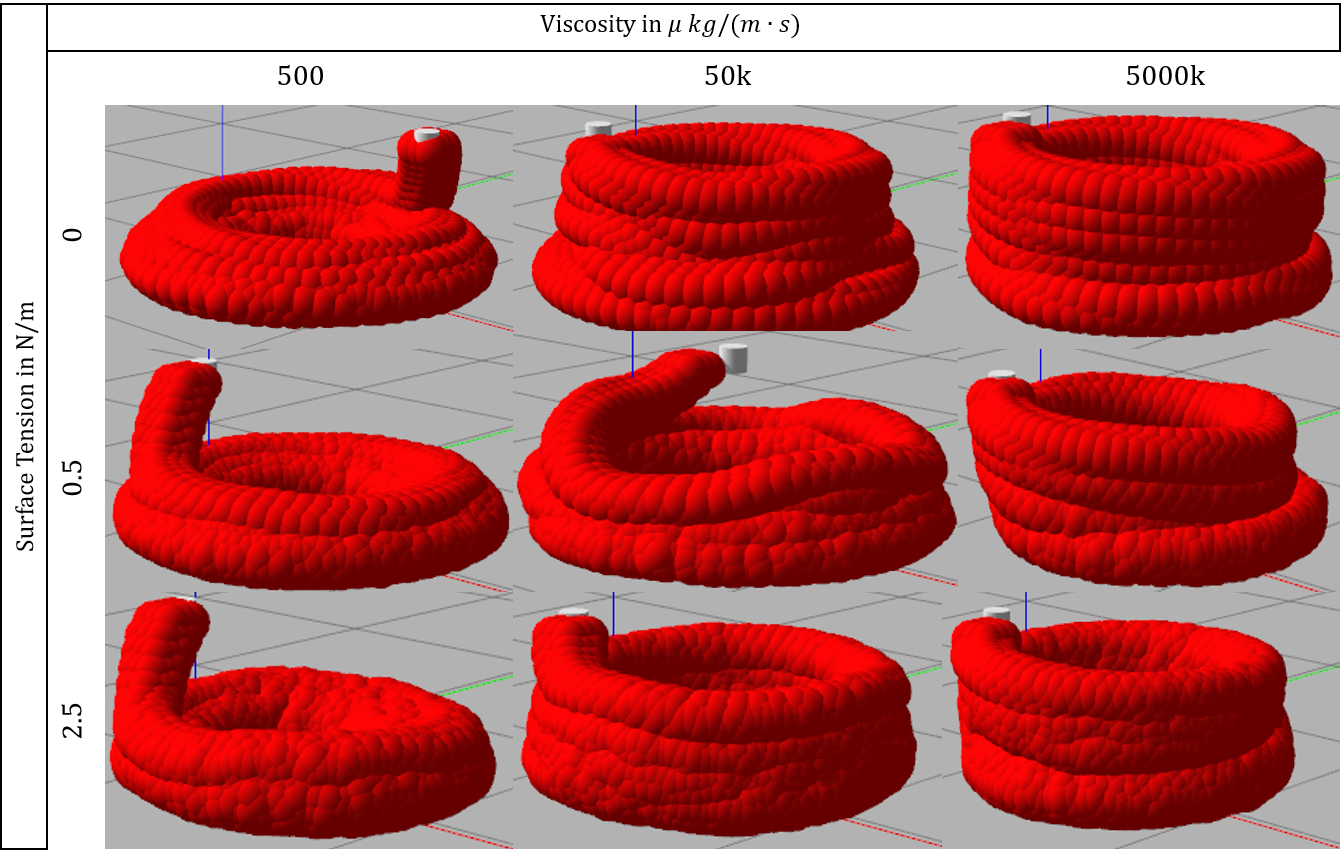}
    \caption{Effect of different surface tension and viscosity.}
    \vspace{-3mm}
    \label{matrix}
\end{figure}

Figure \ref{matrix} shows the capability to simulate a wide array of materials. In addition to surface tension and viscosity, the SPH libraries account for elasticity, vorticity and drag force.

\subsection{Meshing of the simulated particles}

\begin{figure}[h!]
\vspace{-3mm}
    \centering
    \includegraphics[width=0.6\columnwidth]{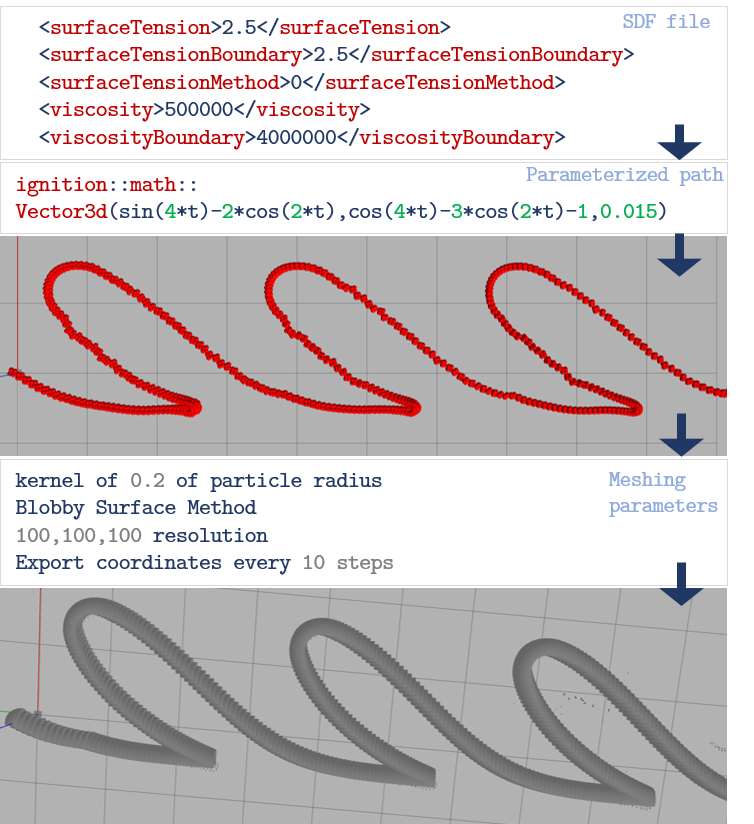}
    \vspace{-3mm}
    \caption{With the given initial conditions, the emitter follows a parameterized path and extrudes the particles that are meshed.}
    \label{mesh_example}
    \vspace{-4mm}
\end{figure}

The coordinates of the center of mass of each rendered entity (e.g. spheres) are passed to the meshing function with a user-adjustable period. All the emitted particles at a given step are used to generate the mesh which is then fed back into the simulation environment through an internally generated SDF file. Figure \ref{mesh_example} shows the process for one layer. 

\subsection{Integration with Mobile Robotics Simulation}

The particle-emitting cylinder is replaced by a moving robot model by attaching the emitter directly to the robot's arm in its description file. A Turtlebot Waffle Pi with openManipulator is used as the mobile printing platform in the Gazebo environment (shown in Figure \ref{teaser}). OpenManipulator is controlled using the \textit{movegroup} interface of the Moveit! package. The Turtlebot itself is controlled via odometry inputs to ROS. The Universal Robot Arm 10 (ur10) was also used to demonstrate the printing of the material by following pre-determined waypoints.

\begin{figure}[h!]
    \centering
    \includegraphics[width=0.4\columnwidth]{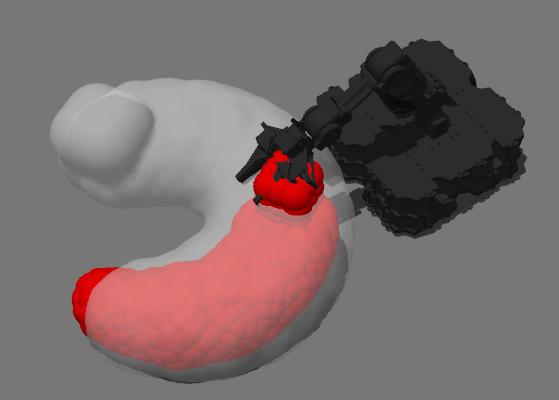} \includegraphics[width=0.5\columnwidth]{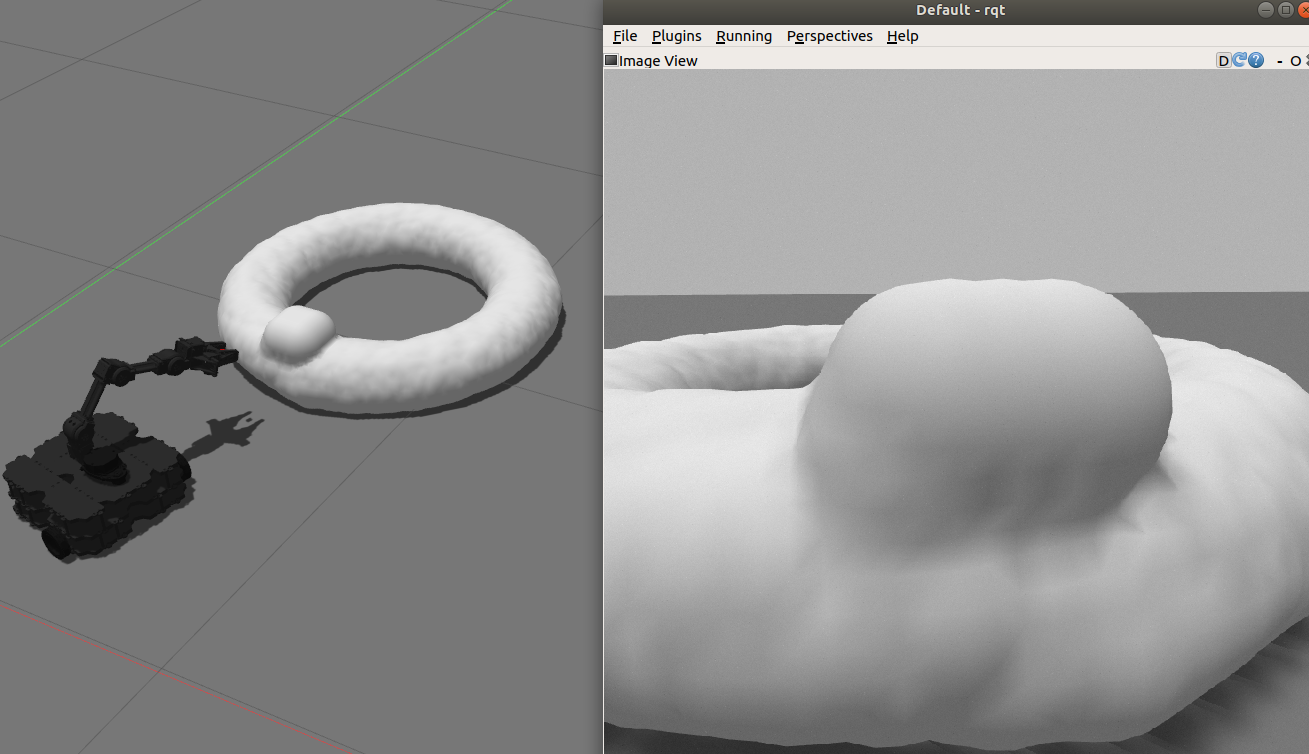}
    \vspace{-2mm}
    \caption{Left: mobile 3D printing TurtleBot following a pre-planned circular trajectory. During printing, the emitted particles are meshed in given time intervals. Particles and meshes can co-exist in the simulation. Right: \textit{robot perception simulation} of mobile 3D printing. The emitted particles are meshed using spherical method during the printing. The printing result is observed by a simulated camera on the mobile robot.}
    \label{robots_example}
    \vspace{-5mm}
\end{figure}

Figure~\ref{robots_example} left shows the printing process for the TurtleBot, where particles are being continuously meshed during run time. Figure~\ref{robots_example} right shows the result of meshing a layer and starting the second while the robot mounted camera perceives the print. This showcases the connection of the material simulation, meshing pipeline and the mobile robot control for mobile 3D printing.

\subsection{Multiple Robot Printing}

\begin{figure}[h!]
    \centering
    \includegraphics[width=0.6\columnwidth]{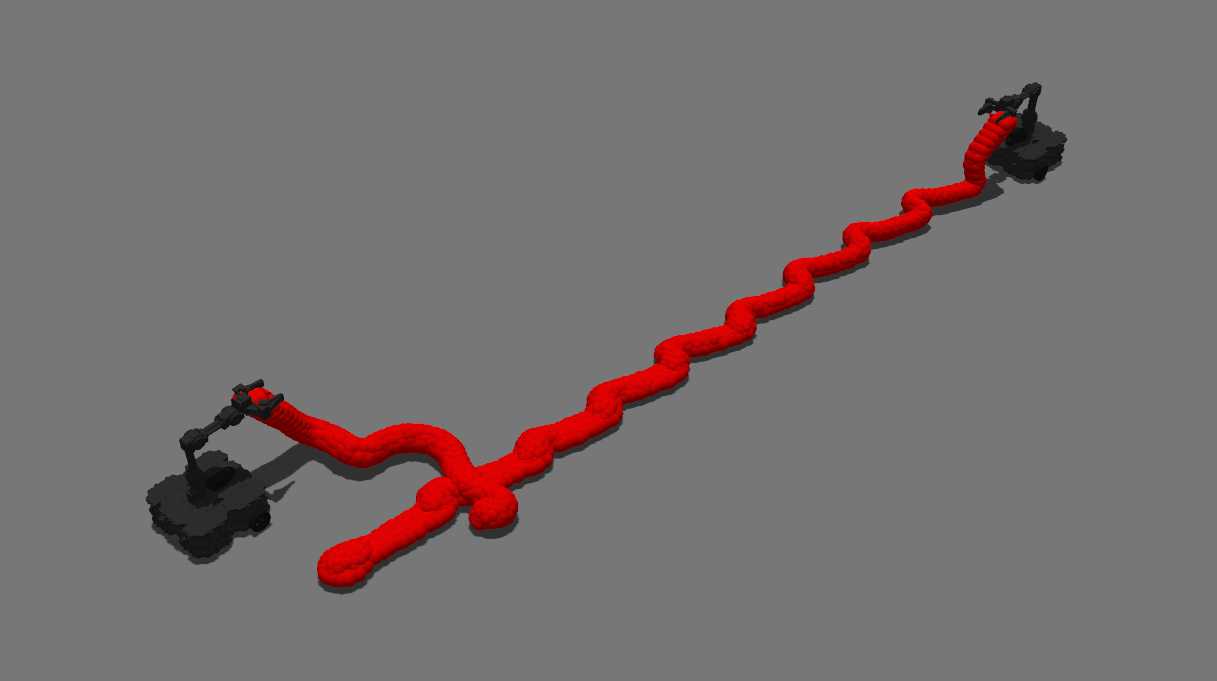}
    \caption{Multiple TurtleBots collaborating through the proposed framework.}
    \label{multi}
    \vspace{-4mm}
\end{figure}

In order for the mobile 3D printing simulation tool to be effective, the above demonstration should be valid for multiple mobile robots. Figure~\ref{multi} shows multiple robots working collaboratively to print a structure. As demonstrated in the experiments above, each robot is able to sense the other robots and the printed material for collision avoidance between the robots themselves and the printed mesh.

\subsection{Performance Analysis}
\begin{figure}[h!]
\vspace{-3mm}
    \begin{center}
    \includegraphics[width=0.9\columnwidth]{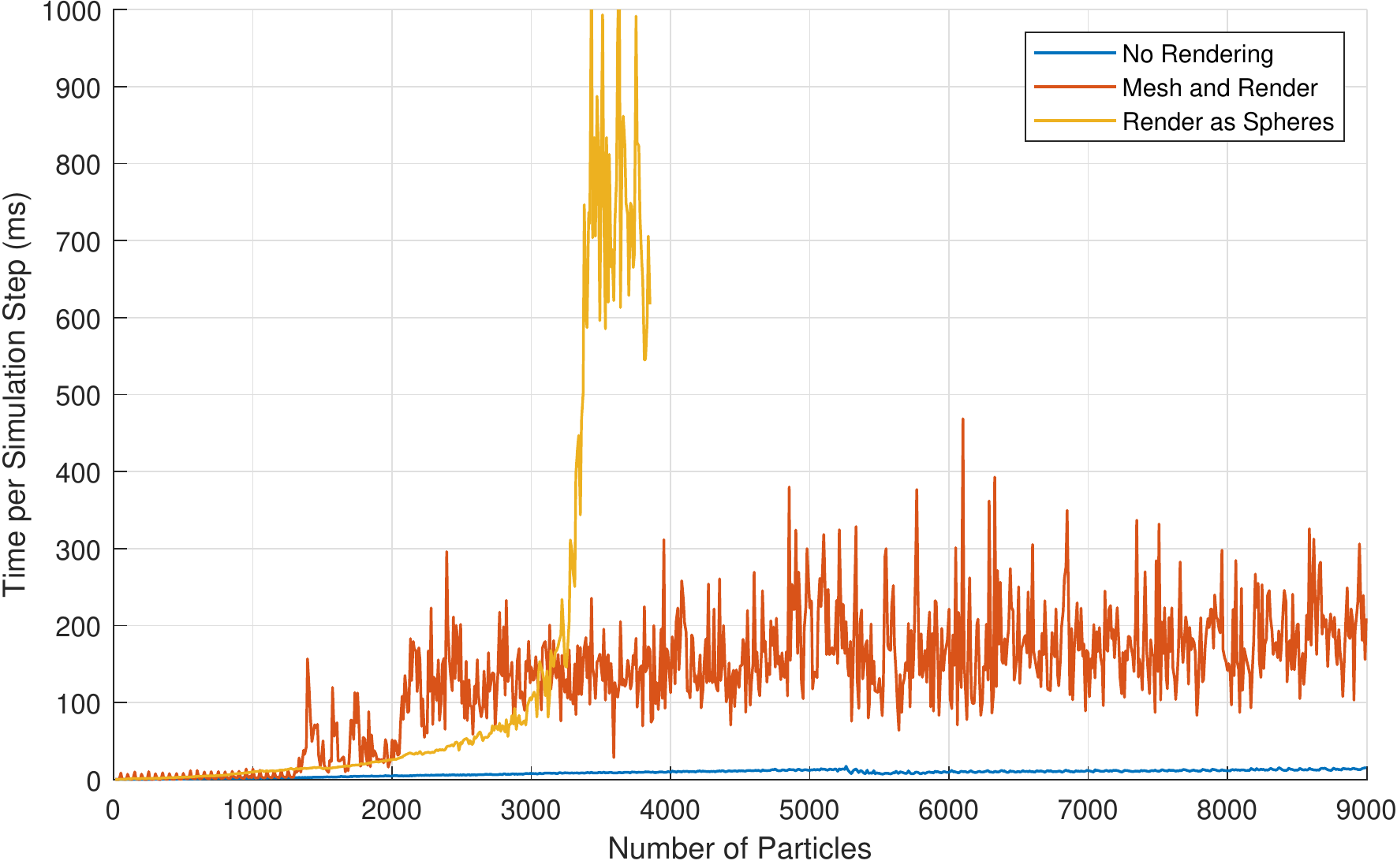}
    \vspace{-3mm}
    \caption{Performance chart comparing run time for different options.}
    \label{fig:performance}
    \vspace{-9mm}
    \end{center}
\end{figure}

Time taken per simulation step was measured over the number of particles emitted using the {\fontfamily{lmtt}\selectfont ctime} library. A baseline where particles are being simulated in the back-end without any visualization is taken as a reference. A significant increase in time per simulation step -- from less than 100ms to near 800ms -- occurs when the total number of particles rendered as spheres exceed 3000, resulting in an almost frozen simulation. The workaround for such phenomenon is for the particles to be  meshed, and then imported into the gazebo GUI as \textit{.obj} models. This method mitigates the steep increase in simulation time, thus achieving stable simulation time around 200ms per step while still running the layer-by-layer simulation as shown in Figure \ref{fig:performance}.

\section{Conclusions}
This work introduced a novel simulation approach for mobile 3D printing in response to the most recent trends in mobile AM robotics and simulation. The versatility of Smooth Particle Hydrodynamics for high viscosity materials is combined with the the ubiquitous dynamic environment of Gazebo. This new approach to simulation solves the computationally costly bottlenecks of mesh-based simulation by adding an external meshing module which lets the user choose the implicit surface generation method and the resolution. The mesh along with the SPH computations are available as rigid entities in the Gazebo multi-robot environment. 

The validity of the proposed method is shown by connecting the mathematical theory of SPH with the explicit meshing principles in a realistic physics simulation model. The attachment of the simulation plugin to different robots is shown along with their capability to sense and interact with these newly generated entities in the scene.

Despite the promising results, the scalability of the Gazebo GUI for a number of particles above the $4^{th}$ order of magnitude requires further work. The most time-stable solution thus far has proven to be the meshing of the first few thousand of particles with the meshing pipeline described in this paper and then reusing the particles or simply using the point coordinates for continuous meshing.

The authors hope that this approach will be used by researcher to help with planning more aspects of their mobile 3D printing processes and inspire people who lack access to resources to simulate and visualize a vast array of printing materials in robotic additive manufacturing processes. 

Future work could include the increase of the Gazebo GUI support for a larger number of rigid {\fontfamily{lmtt}\selectfont entities} than already demonstrated and the full implementation of Material Point Methods which can replace the SPH libraries in the presented workflow and extend the versatility of this tool.

\vspace{-3mm}
\small{
\section*{Acknowledgments}
\vspace{-3mm}

The research is supported by NSF CPS program under CMMI-1932187. Emmanouil Angelidis is supported by the European Union’s Horizon 2020 research and innovation programme under grant agreement No. 945539 (SGA3) Human Brain Project.
}
\vspace{-3mm}


\addtolength{\textheight}{-0.2cm}   


{\small
\bibliographystyle{IEEEtranN}

\bibliography{BibFiles/M3DP,BibFiles/R_S,BibFiles/Material_sim,BibFiles/Print_sim,ai4ce-tpl}
}

\end{document}